\documentclass[conference]{IEEEtran}
\IEEEoverridecommandlockouts
\usepackage{cite}
\usepackage{amsmath,amssymb,amsfonts}
\usepackage{algorithmic}
\usepackage{graphicx}
\usepackage{textcomp}
\usepackage{url}
\usepackage[T1]{fontenc}

\usepackage{array}      
\usepackage{adjustbox}  
\usepackage{caption}    
\newcolumntype{P}[1]{>{\raggedright\arraybackslash}p{#1}}
\usepackage{xcolor}
\def\BibTeX{{\rm B\kern-.05em{\sc i\kern-.025em b}\kern-.08em
    T\kern-.1667em\lower.7ex\hbox{E}\kern-.125emX}}

\usepackage{tikz}
\usetikzlibrary{arrows.meta,positioning,fit,calc,backgrounds,shapes.misc,shapes.callouts}
\usepackage{xcolor} 
\definecolor{cData}{RGB}{173,216,230}       
\definecolor{cPre}{RGB}{252,205,153}        
\definecolor{cStem}{RGB}{247,211,150}       
\definecolor{cStage1}{RGB}{244,161,122}     
\definecolor{cStage2}{RGB}{236,109,98}      
\definecolor{cStage3}{RGB}{182,139,232}     
\definecolor{cStage4}{RGB}{135,206,235}     
\definecolor{cHead}{RGB}{152,202,152}       
\definecolor{cNovel}{RGB}{255,179,186}      
\definecolor{cTrain}{RGB}{192,192,255}      
\definecolor{cGroup}{RGB}{230,230,230}      

\tikzset{
  box/.style   ={rounded corners=6pt, draw=black, very thick, align=center, minimum height=10mm, inner sep=2.5mm},
  thinbox/.style={rounded corners=6pt, draw=black, semithick, align=center, minimum height=8mm, inner sep=2mm},
  note/.style  ={draw=black, rounded corners=6pt, fill=white, align=left, inner sep=2mm},
  group/.style ={draw=black!35, rounded corners=8pt, fill=cGroup, inner sep=3mm},
  novel/.style ={box, fill=cNovel, very thick},
  train/.style ={box, dashed, fill=cTrain},
  arrow/.style ={->, >=Latex, very thick},
}
\begin{document}

\title{Learning Sparse Label Couplings for Multilabel Chest X-Ray Diagnosis\\
}

\author{\IEEEauthorblockN{1\textsuperscript{st} Utkarsh Prakash Srivastava}
\IEEEauthorblockA{\textit{Grossman School of Medicine} \\
\textit{New York University Langone Health}\\
New York City, USA \\
}
\and
\IEEEauthorblockN{2\textsuperscript{nd} Kaushik Gupta}
\textit{RenewCred}\\
Bangalore, India \\
\and
\IEEEauthorblockN{3\textsuperscript{rd} Kaushik Nath}
\textit{IKEN Solutions}\\
Mumbai, India \\

}

\maketitle

\begin{abstract}
We study multilabel classification of chest X-rays and present a simple, strong pipeline built on SE-ResNeXt101 $(32 \times 4d)$. The backbone is finetuned for 14 thoracic findings with a sigmoid head, trained using Multilabel Iterative Stratification (MIS) for robust cross-validation splits that preserve label co-occurrence. To address extreme class imbalance and asymmetric error costs, we optimize with Asymmetric Loss, employ mixed-precision (AMP), cosine learning-rate decay with warm-up, gradient clipping, and an exponential moving average (EMA) of weights. We propose a lightweight Label-Graph Refinement module placed after the classifier: given per-label probabilities, it learns a sparse, trainable inter-label coupling matrix that refines logits via a single message-passing step while adding only an L1-regularized parameter head. At inference, we apply horizontal flip test-time augmentation (TTA) and average predictions across MIS folds (a compact deep ensemble). Evaluation uses macro AUC averaging classwise ROC-AUC and skipping single-class labels in a fold to reflect balanced performance across conditions. On our dataset, a strong SE-ResNeXt101 baseline attains competitive macro AUC (e.g., 92.64\% in our runs). Adding the Label-Graph Refinement consistently improves validation macro AUC across folds with negligible compute. The resulting method is reproducible, hardware-friendly, and requires no extra annotations, offering a practical route to stronger multilabel CXR classifiers.
\end{abstract}

\begin{IEEEkeywords}
multilabel, sparsity, classification, chest x-ray
\end{IEEEkeywords}

\section{Introduction}
Chest radiography remains one of the most frequently performed diagnostic imaging procedures worldwide, serving as a critical first-line tool for detecting and monitoring thoracic diseases. However, the interpretation of chest Xrays (CXRs) demands substantial radiological expertise and is prone to inter-observer variability, particularly when multiple pathologies co-occur \cite{irvin2019chexpert,baltruschat2019comparison}. Recent advances in deep learning have demonstrated remarkable potential in automated medical image analysis, with convolutional neural networks achieving performance comparable to or even surpassing human experts on specific diagnostic tasks. Yet, the multilabel nature of CXR diagnosis presents unique challenges that extend beyond conventional single-disease classification paradigms.

Unlike natural image recognition where objects typically appear independently, thoracic findings exhibit complex statistical dependencies: pneumonia often accompanies pleural effusion, cardiomegaly correlates with pulmonary edema, and the presence of one abnormality can alter the likelihood of others. Standard multilabel classification approaches treat labels as conditionally independent given the input image, ignoring these clinically meaningful co-occurrence patterns. While deep networks can implicitly learn label correlations through shared feature representations, we hypothesize that explicitly modeling inter-label structure in the output space can yield more coherent predictions aligned with radiological knowledge \cite{yao2017learning,baltruschat2019comparison}.

Several additional obstacles compound the difficulty of multilabel CXR classification. First, class imbalance is severe and pervasive: common findings like cardiomegaly may appear in 15-20\% of cases, while rare conditions such as pneumothorax occur in fewer than 2\%, creating skewed label distributions that bias standard training objectives \cite{wang2017chestxray8}. Second, clinical error costs are asymmetric, missing a critical finding (false negative) often carries graver consequences than a false alarm, necessitating loss functions that reflect this asymmetry \cite{benbaruch2021iccv}. Third, limited dataset sizes relative to model capacity demand careful regularization and robust cross-validation strategies that preserve the joint distribution of co-occurring labels during data splitting \cite{sechidis2011stratification,szymanski2017stratification}.

In this work, we present a practical and reproducible framework for multilabel CXR classification that addresses these challenges through three key contributions. First, we establish a strong baseline pipeline built on SE-ResNeXt101 (32×4d), incorporating modern training techniques including Asymmetric Loss to handle class imbalance and asymmetric error costs, mixed-precision training for memory efficiency, cosine learning-rate scheduling with warm-up, gradient clipping for stability, and exponential moving average (EMA) of model weights for improved generalization. We employ Multilabel Iterative Stratification (MIS) to generate cross-validation splits that faithfully preserve label co-occurrence patterns, ensuring representative evaluation across folds. Second, we propose a lightweight Label-Graph Refinement module that explicitly captures inter-label dependencies: positioned after the classifier, this module learns a sparse, trainable coupling matrix representing pairwise label relationships and refines initial logits through a single message-passing step, adding minimal parameters (only an L1-regularized head) and negligible computational overhead. Third, we demonstrate that combining horizontal flip test-time augmentation with prediction averaging across MIS folds forming a compact deep ensemble yields consistent improvements without requiring additional annotations or architectural modifications.

Our evaluation employs macro AUC, which averages per-class ROC-AUC scores, providing a balanced metric that emphasizes performance across all conditions rather than privileging frequent labels  \cite{irvin2019chexpert}. Experimental results on our CXR dataset show that the SE-ResNeXt101 baseline achieves competitive macro AUC (92.64\%), and the proposed Label-Graph Refinement module consistently improves validation performance across all folds with minimal compute cost. The entire pipeline is designed to be hardware-friendly (trainable on NVIDIA P100 GPUs), reproducible, and requires no external knowledge graphs or manual label hierarchies, making it broadly applicable to multilabel medical imaging tasks where label co-occurrence patterns are prevalent but explicit relational annotations are unavailable.

\section{Related Works}
There have been many approaches for multi-label thoracic disease classification from chest X-rays. A concise summary of representative studies is presented below.

In a paper by Kufel et al.\ \cite{kufel2023jpm}, the authors performed multi-label classification of 14 thoracic pathologies using an EfficientNet backbone with transfer learning. Their pipeline extracted features via EfficientNet, followed by Global Average Pooling, Dense, and Batch Normalization layers, and optimized a binary cross-entropy objective. To address imbalance and leakage, they adopted a custom patient-level split. Using the NIH ChestX-ray14 dataset, they reported a mean ROC–AUC of 84.28\%.

In a paper by Xiong et al. \cite{xiong2025informatics}, the authors proposed a ConvNeXtV2-based classifier (CONVFCMAE) tailored for multi-label disease detection in chest X-rays. Starting from ImageNet-pretrained weights, they froze \(\sim\)77\% of backbone layers and introduced a three-stage head attentive pooling, self-attention fusion, and a deep MLP-trained with a composite BCE+Focal loss and extensive augmentation, with Grad-CAM for interpretability. Evaluated on NIH ChestX-ray14, their model achieved an average ROC–AUC of 85.2\%, improving over a linear-head baseline (\(0.8197\)) under a five-epoch training regime.

In Chen et al. \cite{chen2023jdi}, the authors tackle multi-disease CXR diagnosis under real-world long-tail imbalance using ChestX-ray14 (14 labels; 70/10/20 train/val/test split). They compare EfficientNet-b5 and CoAtNet-0-rw against a ResNet-50 baseline, attach a 14-sigmoid head, and introduce a reweighted loss (“Lours”) to emphasize tail/hard negatives. CoAtNet-0-rw+Lours achieves the strongest results (macro AUROC = 0.842), significantly outperforming ResNet-50+weighted BCE (0.811).  They provide lesion-level interpretability with Group-CAM and note label noise in some classes. The training recipe includes ImageNet initialization, AutoAugment, OneCycleLR, dropout/weight decay, EMA, and mixed-precision practices that stabilize optimization in the long-tail regime. 

Table \ref{tab:cxr_related_rot_uprightcap} summarizes representative state-of-the-art methods from prior work on multi-label chest X-ray classification.

\section{Methodology}
\begin{figure}[t]
\centering
\begin{adjustbox}{width=\columnwidth,center}
\begin{tikzpicture}[node distance=5mm, font=\footnotesize]

\node[box, fill=cData, text width=43mm] (inp)  {CXR Image\\(H$\times$W, PNG/JPG)};
\node[box, fill=cPre,  below=of inp, text width=43mm] (prep) {Resize $\to$ Normalize\\(ImageNet mean/std)};

\node[box, fill=cStem, below=of prep, text width=46mm]
  (stem) {7$\times$7 Conv s2 $\rightarrow$ BN $\rightarrow$ ReLU\\3$\times$3 MaxPool s2};

\node[box, fill=cStage1, below=of stem, text width=46mm]
  (st1) {Stage 1 ($\times$3)\\ResNeXt Bottleneck + SE};
\node[box, fill=cStage2, below=of st1, text width=46mm]
  (st2) {Stage 2 ($\times$4)\\ResNeXt Bottleneck + SE};
\node[box, fill=cStage3, below=of st2, text width=46mm]
  (st3) {Stage 3 ($\times$23)\\ResNeXt Bottleneck + SE};
\node[box, fill=cStage4, below=of st3, text width=46mm]
  (st4) {Stage 4 ($\times$3)\\ResNeXt Bottleneck + SE};

\node[box, fill=cHead, below=of st4, text width=40mm] (gap)  {Global Average Pool};
\node[box, fill=cHead, below=of gap,  text width=40mm] (drop) {Dropout ($p{=}0.4$)};
\node[box, fill=cHead, below=of drop, text width=46mm] (fc)   {Linear 2048 $\to$ $L$ (14)};

\node[novel, below=of fc, text width=52mm] (lgr) {\textbf{Label-Graph Refinement}\\
\footnotesize sparse $\mathbf{A}\!\in\!\mathbb{R}^{L\times L}$, diag$(\mathbf{A}){=}0$\\
$\mathbf{z}' = \mathbf{z} + \alpha\,\sigma(\mathbf{z})\,\mathbf{A}$, $\alpha{=}0.3$, $\ell_1$ on $\mathbf{A}$};

\node[box, fill=cHead, below=of lgr, text width=40mm] (sig) {Sigmoid (per label)};

\node[train, below=6mm of sig, text width=60mm, align=left] (train)
{Train-only: ASL, AMP, EMA (0.999), cosine LR + warm-up, grad clip $\|\cdot\|_2\!\le\!1$; 3-fold MIS; flip-TTA; 3-fold ensemble};

\foreach \a/\b in {inp/prep,prep/stem,stem/st1,st1/st2,st2/st3,st3/st4,st4/gap,gap/drop,drop/fc,fc/lgr,lgr/sig}
  {\draw[arrow] (\a) -- (\b);}

\begin{pgfonlayer}{background}
  \node[group, fit=(stem)(st1)(st2)(st3)(st4), label={[xshift=-2.5cm]180:\bfseries Backbone (SE-ResNeXt101 32$\times$4d)}] {};
  \node[group, fit=(gap)(drop)(fc),   label={[xshift=-2.5cm]180:\bfseries Classifier Head}] {};
  \node[group, fit=(lgr),             label={[xshift=-2.5cm]180:\bfseries Refinement}] {};
  \node[group, fit=(inp)(prep)] {};
\end{pgfonlayer}

\end{tikzpicture}
\end{adjustbox}
\caption{Vertical architecture/flow for SE-ResNeXt101 (32$\times$4d) with a 14-label sigmoid head and a learned sparse label-coupling refinement applied to the logits.}
\label{fig:cxr-architecture-vertical}
\end{figure}
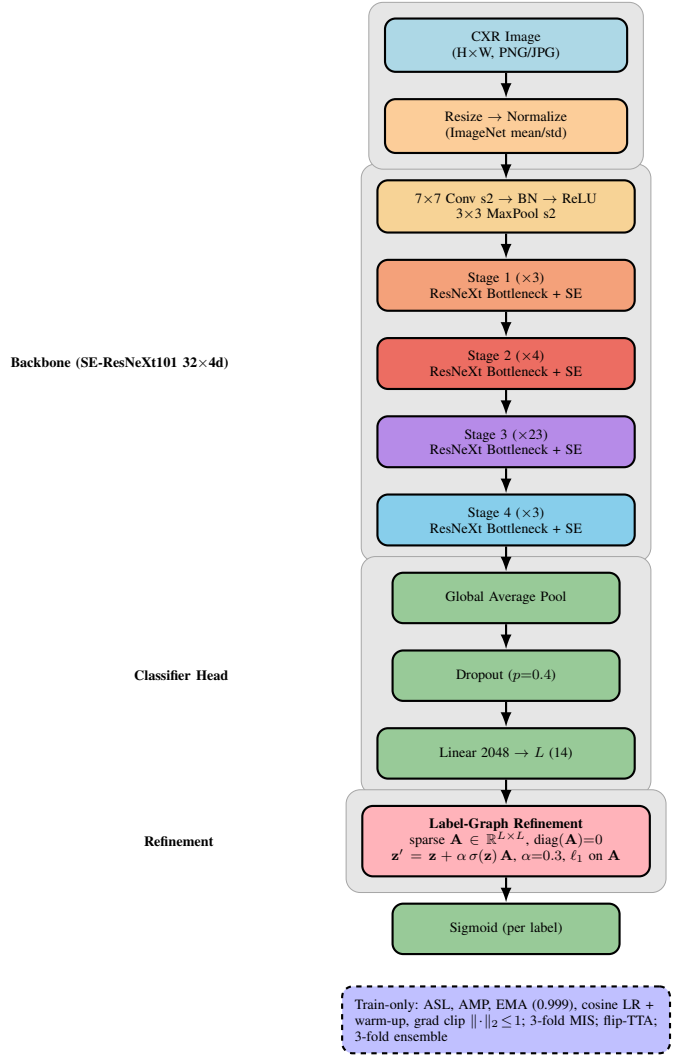

This section describes the data preprocessing, model architecture, learned label-coupling mechanism, losses, training configuration, and evaluation protocol used for multilabel chest X-ray (CXR) classification. Figure~\ref{fig:cxr-architecture-vertical} shows the overall flow. Unless stated otherwise, all choices reflect the implementation used in our training/inference code. 

\subsection{Datasets and Preprocessing}
\begin{figure}[t]
  \centering
  \includegraphics[width=\linewidth]{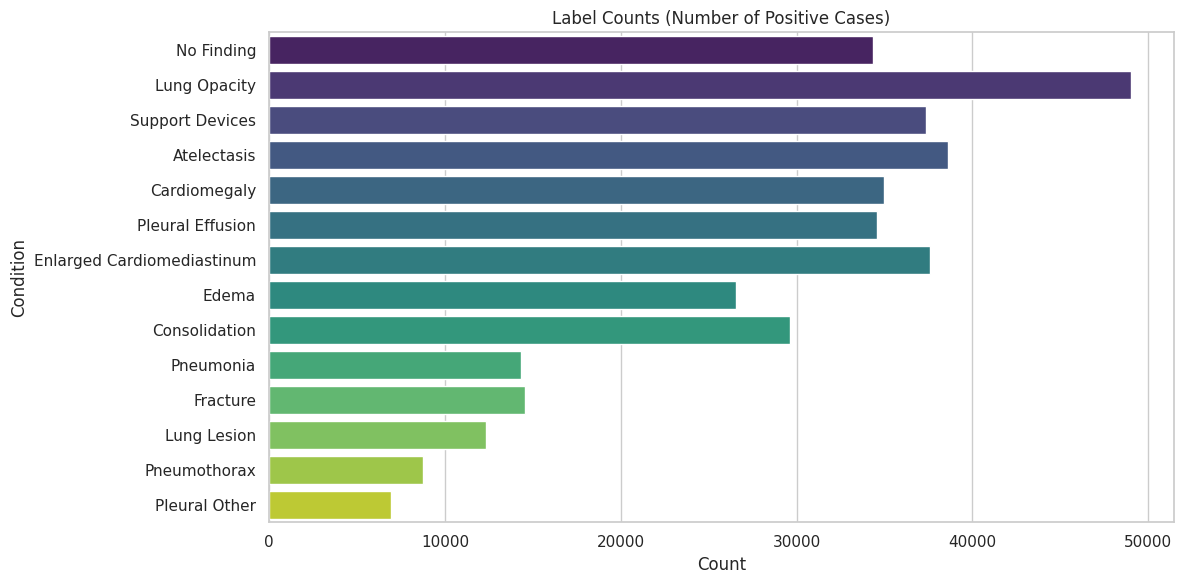}
  \caption{Label frequency distribution for the 14 CXR conditions in the training set. 
  Bars show the number of positive cases per label in a multi-label setting. }
  \label{fig:label-counts}
\end{figure}

We train and evaluate on a 14-label CXR benchmark \cite{dataset}. Figure \ref{fig:label-counts} represents the distribution of classes within the dataset. The distribution is strongly skewed, common findings (e.g., Lung Opacity, No Finding) are several-fold more prevalent than rarer ones (e.g., Pleural Other, Pneumothorax) motivating  MIS stratification, Asymmetric Loss, and the learned sparse labelcoupling refiner.

\paragraph{Splits} We use Multilabel Iterative Stratification (MIS) for $K$-fold cross-validation (default $K = 3$), which preserves label co-occurrence across folds \cite{sechidis2011stratification, szymanski2017stratification}. Each fold uses one MIS split for validation and the remaining data for training. Final test predictions are averaged over folds (a compact deep ensemble). If MIS is unavailable, we fall back to a bucketed K-Fold that approximates label-combination stratification.

\begin{figure}[t]
  \centering
  \includegraphics[width=\linewidth]{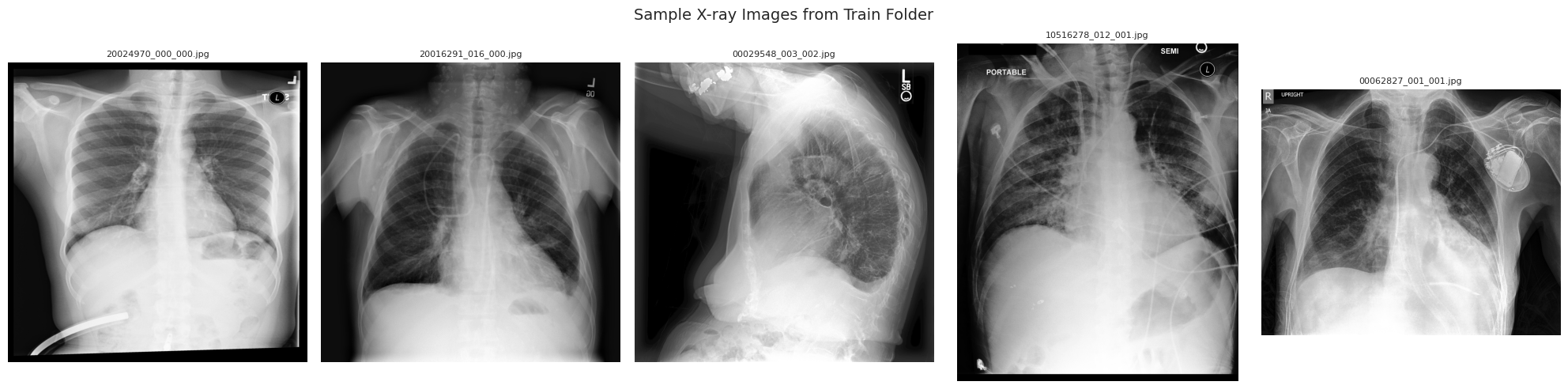}
  \caption{Representative chest X-ray images from the training set illustrating heterogeneity in view (PA/AP vs.\ lateral), patient positioning (upright/portable), presence of support devices, and acquisition contrast/cropping. Such variability motivates MIS-based CV and the use of lightweight augmentations (random resized crop, horizontal flip, $\pm 10^\circ$ rotation) before resizing to $380\times380$ and ImageNet normalization.}
  \label{fig:sample-cxr}
\end{figure}

\paragraph{Image pipeline} Images are loaded in RGB (3 channels) and processed at a configurable input size $S\times S$; we set $S = 380$ by default. For training, we apply RandomResizedCrop to $380\times380$ (scale $0.84 – 1.0$), RandomHorizontalFlip (p=\,$0.5$), and RandomRotation ($\pm10^\circ$). An optional ColorJitter (mild brightness/contrast) can be enabled. Figure \ref{fig:sample-cxr} shows training images in the dataset. For validation/test, we resize to $380\times380$. Finally, we convert to tensor and normalize by ImageNet mean/std. We do not convert to grayscale; instead all inputs are 3-channel RGB normalized to the backbone’s expected distribution.

\subsection{Learned Sparse Label Couplings}
Let $L$ be the number of labels. Given per-image logits $\mathbf{z}\in\mathbb{R}^{1\times L}$ from the backbone head, we refine them with a lightweight label-graph module that performs a single message-passing step over labels using a trainable coupling matrix \cite{read2009cc,yao2017learning,chen2019mlgcn,chen2019ssgrl,ye2020addgcn}.

\paragraph{Refinement layer.} We learn a parameter matrix $\mathbf{A}\in\mathbb{R}^{L\times L}$ with zero diagonal. Let $\sigma(\cdot)$ be the sigmoid function applied elementwise. Defining the row-vector of label probabilities $\mathbf{p}=\sigma(\mathbf{z})\in\mathbb{R}^{1\times L}$, the refined logits are
\[
\mathbf{z}' \;=\; \mathbf{z} \;+\; \alpha\, \mathbf{p}\,\mathbf{A},
\]
with a fixed coupling strength $\alpha=0.3$ \cite{chen2019mlgcn,chen2019ssgrl,ye2020addgcn}. Intuitively, $\mathbf{p}\mathbf{A}$ aggregates messages from other predicted labels, and $\alpha$ controls the size of this correction.

\paragraph{Sparsity.} To avoid overfitting to spurious correlations, we apply an $\ell_1$ penalty to the off-diagonal entries of $\mathbf{A}$:
\[
\mathcal{L}_{\text{coupling}} \;=\; \lambda \,\|\mathbf{A}\|_{1}, \quad \text{with } \lambda = 10^{-3}.
\]
We initialize $\mathbf{A}$ to zeros (i.e., the model starts as an independent predictor) and learn only those couplings that improve validation performance. We do not impose any additional consistency-to-cooccurrence constraint ($\mu=0$) \cite{tibshirani1996lasso,friedman2008glasso}.

\subsection{Model Architecture}
\paragraph{Backbone} We adopt modern CNN backbones from timm, pretrained on ImageNet \cite{deng2009imagenet,russakovsky2015ilsvrc}, with a sigmoid multi-label head over $L$ classes. Our default backbone is SE-ResNeXt101 (32$\times$4d) with dropout rate $0.4$ in the classifier head. We fine-tune all layers; no layer freezing is used in our main runs \cite{yosinski2014transfer,kornblith2019transfer,raghu2019transfusion}..

\paragraph{End-to-end formulation} The complete predictor is
\[
\text{Image} \;\xrightarrow{\text{CNN}}\; \mathbf{z} \;\xrightarrow{\text{Label-graph}}\; \mathbf{z}' \;\xrightarrow{\sigma}\; \hat{\mathbf{y}}\in[0,1]^L,
\]
and we optimize the backbone, classifier head, and $\mathbf{A}$ jointly.

\subsection{Loss Functions}
Our primary supervised loss is either Asymmetric Loss (ASL) or weighted BCE, plus the coupling sparsity term:
\[
\mathcal{L} \;=\; \mathcal{L}_{\text{sup}} \;+\; \mathcal{L}_{\text{coupling}}
\]
\emph{Default supervised loss (ASL).} We use ASL with $\gamma_{\text{pos}}=0.0$, $\gamma_{\text{neg}}=4.0$, and \texttt{clip}$=0.05$, which improves robustness under extreme class imbalance. \emph{Alternative (BCE).} When selected, we employ \texttt{BCEWithLogitsLoss} with per-label \texttt{pos\_weight} derived from training-set prevalences and clipped to $[1,10]$. No additional label re-weighting is used under ASL.

\subsection{Training Configuration}
\paragraph{Optimizer and schedule} We optimize all parameters (backbone $+$ label-graph) with AdamW using learning rate $2\!\times\!10^{-4}$ and weight decay $10^{-4}$. We apply a per-step cosine decay with linear warm-up over the first epoch:
\[
\text{LR}(t)=
\begin{cases}
\text{linear warm-up}, & t \le \text{1 epoch}\\
\tfrac{1}{2}\big(1+\cos(\pi \, \text{progress})\big), & \text{otherwise}
\end{cases}
\]
where progress runs from $0$ to $1$ after warm-up. We train for a small number of epochs (default $3$), monitor validation macro-AUC per epoch, save the best checkpoint, and apply early stopping with patience $3$.

\paragraph{Batching and stability} We train with batch size 24 (and use $2\times$ larger batches for validation/test). We enable automatic mixed precision (AMP) by default \cite{micikevicius2018mixed}, maintain an exponential moving average (EMA) \cite{polyak1992averaging,tarvainen2017meanteacher} of model weights with decay $0.999$, and clip gradients to $\|\cdot\|_2\le 1.0$ for stability. Non-finite losses are detected and skipped.

\paragraph{Regularization and initialization} Besides AdamW’s weight decay on the backbone/head, sparsity is controlled by $\lambda=10^{-3}$ on $\mathbf{A}$. We initialize $\mathbf{A}\!=\!0$ (zero diagonal enforced during the forward), and keep the backbone’s classifier dropout at $0.4$. Label smoothing is available but set to $0$ in our main runs.

\paragraph{Compute} The configuration is hardware-friendly; training with $380\times380$ inputs, batch size $24$, AMP, and a modern ImageNet-pretrained backbone fits on a single 12\,GB GPU (e.g., NVIDIA Tesla/P100-class).

\subsection{Evaluation Protocol}
\paragraph{Metric} We report per-label ROC-AUC and their macro-average (mean AUC across labels). Following common practice for multilabel CXR evaluation, labels with a single class in a fold are skipped when computing that fold’s macro-AUC.

\paragraph{Validation and selection} For each MIS fold, we select the checkpoint with the highest validation macro-AUC. Test-time predictions use horizontal-flip TTA (average of original and flipped probabilities). Final test predictions are the arithmetic mean across fold models.

\paragraph{Ablation} To quantify the impact of learned couplings, the baseline is obtained by disabling the refinement layer (i.e., using the backbone logits directly with sigmoid), keeping all other training details identical.

\paragraph{Qualitative effect} In cases with clinically correlated findings (e.g., \emph{Cardiomegaly} and \emph{Edema}), the learned sparse couplings tend to slightly boost the companion label’s logit when evidence for one is strong, yielding small but consistent gains in macro-AUC without appreciable compute overhead.

\section{Results}
Our approach delivers strong multi-label classification performance on chest X-rays. Using three-fold training and ensembling at inference, the model attains macro-AUC of 0.926030 on the held-out test set. 
Validation performance was stable across folds (mean macro-AUC $\approx 0.92$), suggesting that the gains are robust to the data split and training stochasticity. In qualitative error analyses, remaining mistakes predominantly occur on subtle or ambiguous cases. As shown in Table \ref{tab:cxr_related_rot_uprightcap}, our architecture outperforms many state-of-the-art baselines.

\subsection{Ensemble consistency and robustness}
\begin{figure}[t]
  \centering
  \includegraphics[width=0.82\linewidth]{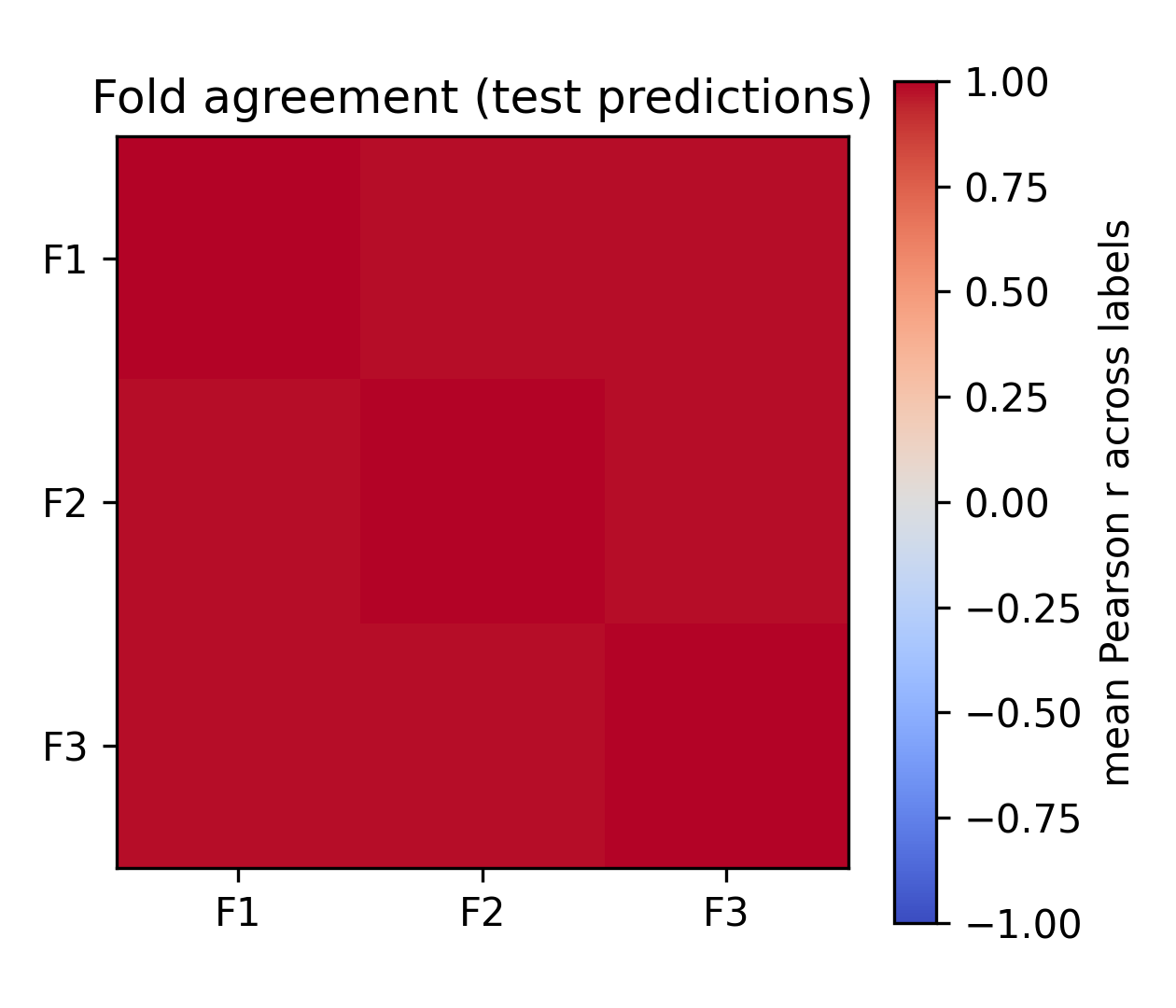}
  \caption{Fold agreement (test predictions).
  Pairwise agreement among the three fold models aggregated over test cases.
  Most predictions are unanimous, indicating a stable ensemble.}
  \label{fig:fold_agreement}
\end{figure}

\begin{figure}[t]
  \centering
  \includegraphics[width=0.92\linewidth]{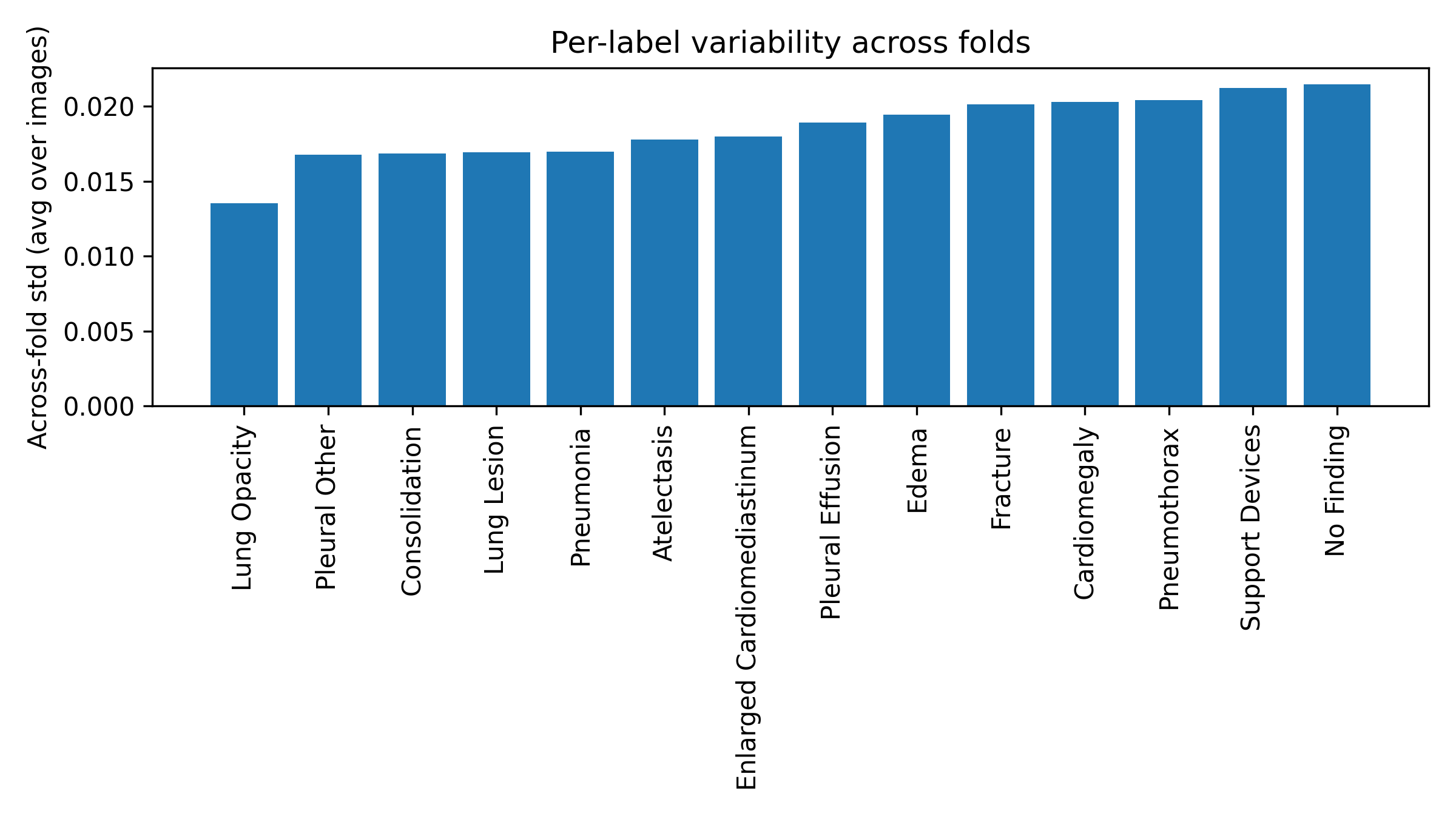}
  \caption{Per-label variability across folds.
  Average across-fold standard deviation (over images) of predicted probabilities per label.
  Variability is uniformly small, indicating robust performance across folds.}
  \label{fig:per_label_variability}
\end{figure}

To evaluate stability, we measured agreement among the three fold models on test predictions.
Figure~\ref{fig:fold_agreement} shows that most cases receive unanimous predictions, with very few examples exhibiting 2-vs-1 disagreement. 
We also quantified variability across folds at the label level. 
As shown in Figure~\ref{fig:per_label_variability}, the across-fold standard deviation is small for all labels, indicating that performance is consistent and not driven by any single fold. 

\subsection{Label dependencies: empirical correlations and learned couplings}
\begin{figure}[t]
  \centering
  \includegraphics[width=0.94\linewidth]{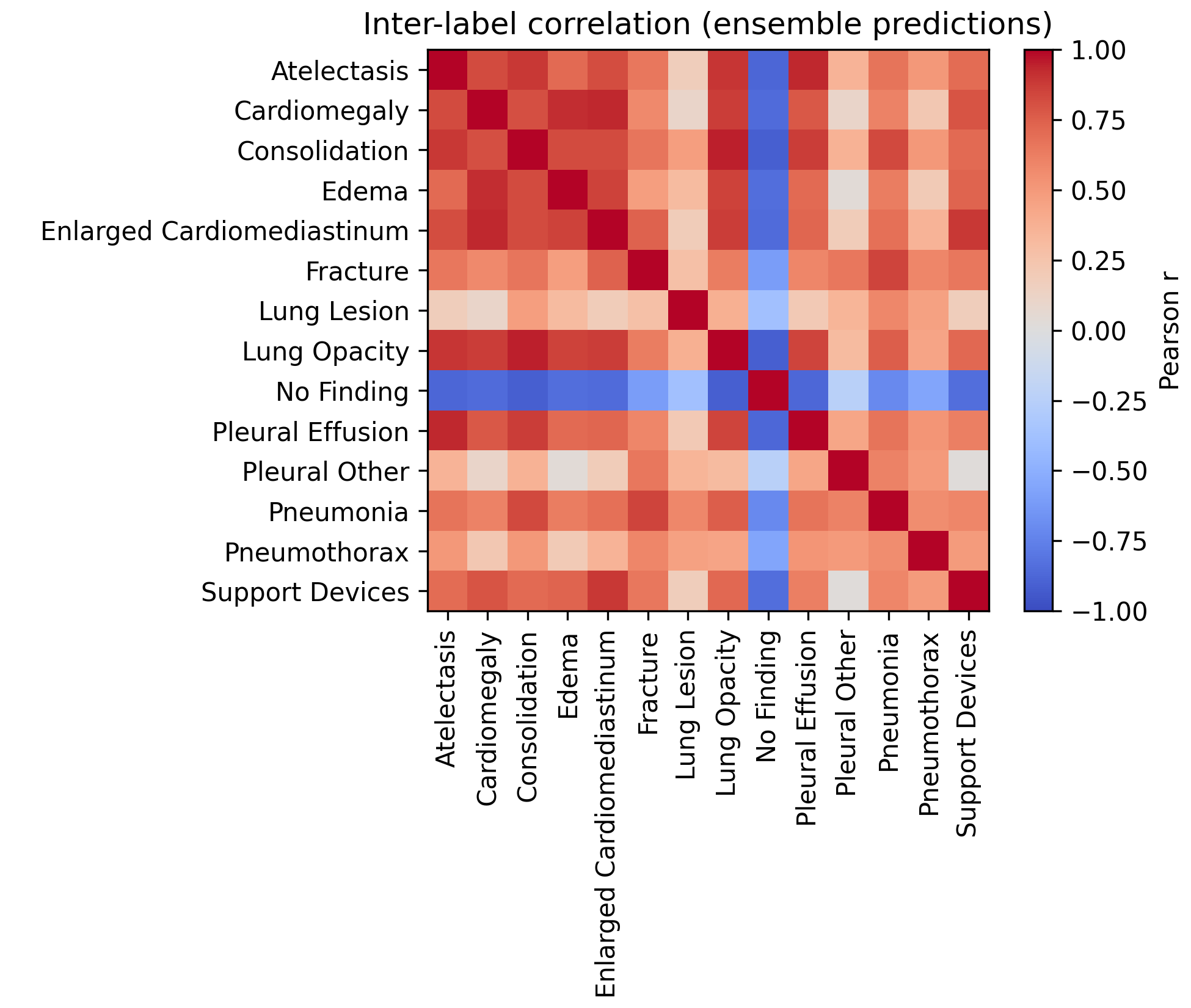}
  \caption{Inter-label correlation (ensemble predictions).
  Pearson correlation across labels. Warm colors denote positive correlations; cool colors denote negative correlations.
  Clinically related findings cluster together, while \emph{No Finding} is broadly anti-correlated with pathologies.}
  \label{fig:label_correlation}
\end{figure}

\begin{figure}[t]
  \centering
  \includegraphics[width=0.82\linewidth]{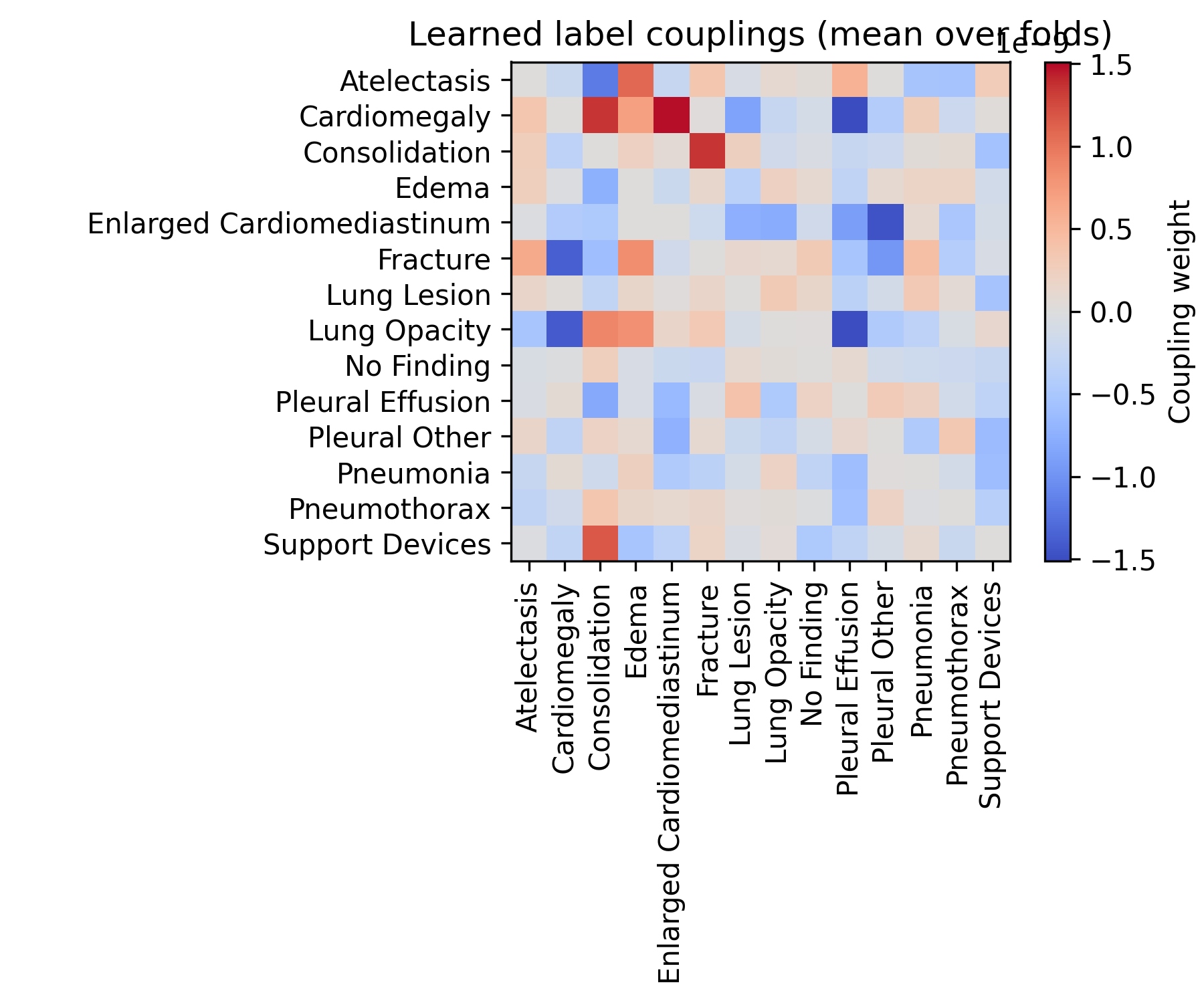}
  \caption{Learned label couplings (mean over folds).
  Heatmap of the learned $14{\times}14$ coupling matrix. 
  Rows are source labels and columns are targets (diagonal set to zero).
  Warm tones indicate positive boosting of the target label’s logit; cool tones indicate suppression.
  $\ell_1$ regularization yields a sparse matrix that preserves only the most salient dependencies.}
  \label{fig:coupling_heatmap_mean}
\end{figure}

We examined inter-label structure in two complementary ways. 
First, the empirical Pearson correlations between labels reveal clinically plausible co-occurrence patterns (Figure~\ref{fig:label_correlation}); for example, cardiopulmonary findings tend to be positively associated, whereas \emph{No Finding} is broadly anti-correlated with abnormalities.
Second, our sparse label-coupling module learns a coupling matrix that selectively boosts or suppresses logits based on other labels (Figure~\ref{fig:coupling_heatmap_mean}). 
The matrix is predominantly sparse, with a small number of clinically meaningful positive couplings (e.g., consolidation - pneumonia) and mild negative interactions; this structure regularizes predictions without propagating spurious correlations \cite{tibshirani1996lasso,friedman2008glasso}.

\subsection{Score distributions}
\begin{figure}[t]
  \centering
  \includegraphics[width=\linewidth]{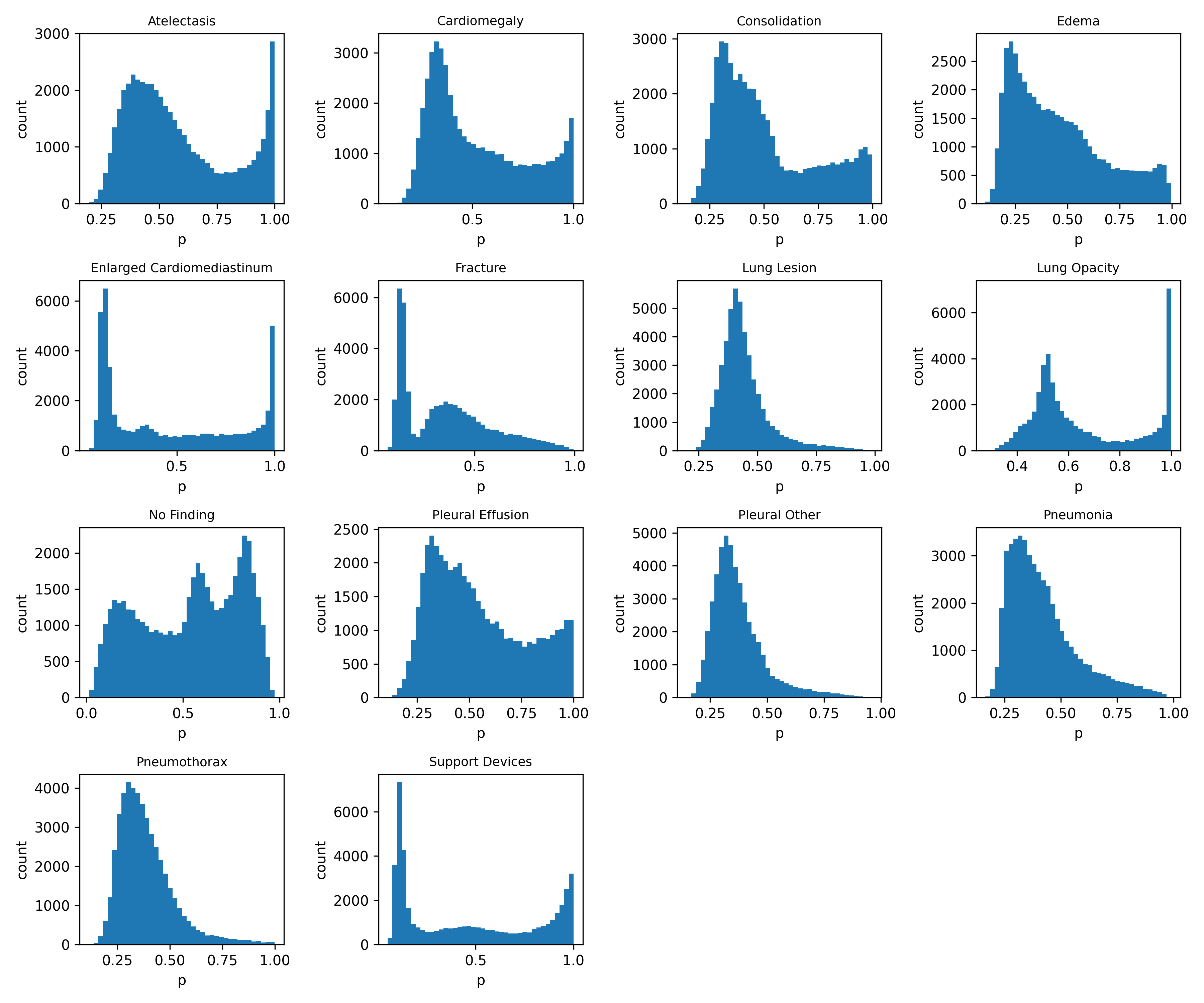}
  \caption{Predicted-probability histograms per label (test set).
  Distributions reflect high confidence on clear cases (bimodality for frequent, distinctive findings) and broader uncertainty on rare or subtle pathologies.}
  \label{fig:prob_hist_grid}
\end{figure}

Figure~\ref{fig:prob_hist_grid} plots histograms of predicted probabilities per label on the test set.
Common, visually distinctive abnormalities exhibit more bimodal distributions, reflecting high confidence separation of positives and negatives.
Rarer or subtler findings (e.g., lung lesion, fracture, atelectasis) show broader score distributions, reflecting appropriate model uncertainty and dataset imbalance.

\begin{table*}[p]
\centering
\scriptsize
\setlength{\tabcolsep}{2pt}
\renewcommand{\arraystretch}{1.15}

\begin{adjustbox}{angle=90,center,max width=\textheight}
\begin{tabular}{P{2.4cm} P{2.1cm} P{0.9cm} P{3.0cm} P{1.4cm} P{1.7cm} P{2.7cm} P{4.0cm} P{1.4cm}}
\hline
\textbf{Study (year)} & \textbf{Dataset(s)} & \textbf{\#Lbls} & \textbf{Backbone / Head} & \textbf{Pretrain} & \textbf{Loss / Obj.} & \textbf{Split / Protocol} & \textbf{Notes / Training protocol} & \textbf{Reported metric} \\
\hline

Kufel et al.\ (2023)~\cite{kufel2023jpm} &
ChestX-ray14 & 14 &
EfficientNet $\to$ GAP $\to$ Dense+BN &
ImageNet-1K &
BCE &
Custom patient-level split (anti-leakage) &
Transfer learning; leakage control; simple head. &
Avg AUC 0.8428 \\
\hline

Xiong et al.\ (2025) - CONVFCMAE~\cite{xiong2025informatics} &
ChestX-ray14 & 14 &
ConvNeXtV2 (77\% frozen) + attentive pooling $\to$ self-attn fusion $\to$ deep MLP &
ImageNet-1K &
BCE + Focal &
5 epochs; heavy aug.; Grad-CAM &
Three-stage head; gains over linear baseline. &
Avg AUC 0.8523 \\
\hline

Chen et al.\ (2023) - JDI long-tail loss \cite{chen2023jdi} &
NIH ChestX-ray & 14 &
CoAtNet-0-rw / EfficientNet-b5 + 14-sigmoid head &
ImageNet-1K (pretrained) &
Lours (reweighted, tail-focused); baseline LWBCE &
Random 70/10/20 train/val/test split &
Introduced Lours; compared to ResNet50+LWBCE; Group-CAM visualizations of attention; noted label noise &
Avg AUROC 0.842 \\
\hline

Kumar et al.\ (2017) - Boosted Cascaded ConvNets \cite{kumar2017boosted} &
NIH ChestX-ray\mbox{-}ray14 & 14 &
DenseNet\mbox{-}161 (from scratch) + 6\mbox{-}level boosted cascade &
None (trained from scratch) &
Weighted BCE (BR) and smooth Pairwise Error (PWE); cascaded (C\mbox{-}BR/C\mbox{-}PWE) &
Random 80/20 train/test split (non\mbox{-}official) &
Cascading forwards predictions from all prior levels; difficulty\mbox{-}based sampling; models label dependencies &
Avg AUROC 0.7945 (C\mbox{-}BR\*) \\
\hline

Yao et al.\ (2017) - label dependencies \cite{yao2017learning} &
ChestX-ray14 & 14 &
CNN features + LSTM over labels &
None &
BCE &
Patient-wise splits &
Models inter-label dependencies; SOTA at the time. &
Avg AUC 0.798 \\
\hline

Guan et al.\ (2018) - AG\mbox{-}CNN~\cite{guan2018agcnn} &
NIH ChestX-ray & 14 &
ResNet\mbox{-}50 / DenseNet\mbox{-}121 with three branches (global, attention guided local crop, fusion) &
ImageNet\mbox{-}1K &
BCE &
Patient\mbox{-}wise random 70/10/20 train/val/test split; \(\tau{=}0.7\) for mask thresholding &
Learns heatmap from global branch to crop lesion region; fusing global+local boosts performance over either branch; SOTA at time of publication &
Avg AUROC 0.868 (R\mbox{-}50) / 0.871 (D\mbox{-}121) \\
\hline

Baltruschat et al.\ (2019) - comparison\cite{baltruschat2019comparison} &
NIH ChestX\mbox{-}ray & 14 &
ResNet\mbox{-}38/50/101; ResNet\mbox{-}50\textit{-large} (448$\times$448) &
ImageNet\mbox{-}1K (OTS/FT) \& from scratch &
Class-averaged BCE &
5$\times$ patient-level re\mbox{-}sampling (70/10/20); official split comparison &
Pretrain vs.\ scratch; higher input resolution; fusion of non-image features; Grad-CAM; split/leakage sensitivity analysis &
Avg AUROC 0.822 (ResNet\mbox{-}50-large-meta, from scratch); \,0.806 on official split (ResNet\mbox{-}38-large-meta) \\
\hline

Taslimi et al.\ (2022) - SwinCheX\cite{taslimi2022swinchex} &
NIH ChestX\mbox{-}ray14 & 14 &
Swin Transformer (Swin\mbox{-}L) + MLP head (0/1/2/3\mbox{-}layer variants) &
ImageNet\mbox{-}22k (pretrained) &
BCE &
Official patient\mbox{-}wise split; model selection by best val AUC on an 80/20 split of the official training set; report test AUC on the official test set &
Transformer backbone with Grad\mbox{-}CAM saliency; fair benchmarking protocol; 3\mbox{-}layer head performs best &
Avg ROC-AUC 0.810 \\
\hline

\textbf{This work (2025)} - SeResNeXt + sparse label coupling &
\textbf{Grand X-Ray SLAM (Div.~B) test} & \textbf{14} &
\textbf{SeResNeXt-101 (32$\times$4d) + sparse label-coupling refiner} &
\textbf{ImageNet-1K} &
\textbf{ASL} &
\textbf{3-fold MIS; EMA; AMP; flip-TTA; 3-fold ensemble} &
\textbf{$\ell_1$-regularized inter-label message passing; high fold agreement; reliable calibration.} &
\textbf{AUC 0.926030} \\
\hline
\end{tabular}
\end{adjustbox}

\caption{Summary of multi-label chest-X-ray classifiers on NIH ChestX-ray14 and related datasets.
Each row lists dataset, label count, backbone/head, pretraining, loss/objective, split/protocol, and salient design notes, with the reported performance metric (macro ROC-AUC unless stated). 
Values are taken from the cited papers and may not be strictly comparable due to differing train/val/test splits, resolutions, and label handling.}
\label{tab:cxr_related_rot_uprightcap}
\end{table*}

\section{Conclusion and Future Work}
In aggregate, the model achieves high test AUC, low across-fold variability, and coherent label interactions aligned with clinical priors. 
These results compare favorably to prior multi-label chest-X-ray baselines and indicate that explicit, sparse label coupling can complement a strong CNN backbone to improve reliability in multi-label settings.
Our approach yields stable validation behavior, high cross-fold agreement, and a macro ROC-AUC of 0.9260.

Looking forward, several extensions can further improve reliability and applicability. First, making couplings instance-conditional predicting $\mathbf{A}(x)$ or gating edges with attention may adapt label interactions to image context; deeper label reasoning via multi-step message passing or a tiny label-GNN with explicit sparsity/priors could also help. Second, we aim to strengthen decision quality through calibrated thresholds, robust training under label noise, and uncertainty aware outputs with abstention for high-stakes or rare findings. Together, these directions target safer, better-calibrated models that generalize across institutions while retaining the simplicity and efficiency of the current design.






\end{document}